\documentclass[sigconf]{acmart}

\usepackage{booktabs} 

\usepackage{booktabs} 

\usepackage{url}
\usepackage{bm}
\usepackage{multirow}
\usepackage{balance}
\setcopyright{rightsretained}

\acmDOI{10.475/123_4}

\acmISBN{123-4567-24-567/08/06}

\acmConference[MM'18]{ACM Multi Media conference}{October 22-26, 2018}{Seoul, Korea}
\acmYear{2018}
\copyrightyear{2018}

\begin{document}
\title{Attentive Crowd Flow Machines}

\author{Lingbo Liu}
\affiliation{%
  \institution{Sun Yat-sen University}
}
\email{liulingb@mail2.sysu.edu.cn}

\author{Ruimao Zhang}
\affiliation{%
  \institution{The Chinese University of Hong Kong}
}
\email{ruimao.zhang@ieee.org}

\author{Jiefeng Peng}
\affiliation{%
  \institution{Sun Yat-sen University}
}
\email{jiefengpeng@gmail.com}

\author{Guanbin Li}
\authornote{Corresponding author is Guanbin Li. This work was supported by the State Key Development Program under Grant 2018YFC0830103, the National Natural Science Foundation of China under Grant 61702565 and Grant 51778033, the Science
and Technology Planning Project of Guangdong Province under Grant 2017B010116001, and was also sponsored by CCF-Tencent Open Research Fund.}
\affiliation{%
  \institution{Sun Yat-sen University}
}
\email{liguanbin@mail.sysu.edu.cn}

\author{Bowen Du}
\affiliation{%
 \institution{Beihang University}
 \institution{BDBC lab}
}
\email{dubowen@buaa.edu.cn}

\author{Liang Lin}
\affiliation{%
  \institution{Sun Yat-sen University}
}
\email{linliang@ieee.org}

\renewcommand{\shortauthors}{Liu et al.}

\begin{abstract}
Traffic flow prediction is crucial for urban traffic management and public safety. Its key challenges lie in how to adaptively integrate the various factors that affect the flow changes. In this paper, we propose a unified neural network module to address this problem, called Attentive Crowd Flow Machine~(ACFM), which is able to infer the evolution of the crowd flow by learning dynamic representations of temporally-varying data with an attention mechanism.
Specifically, the ACFM is composed of two progressive ConvLSTM units connected with a convolutional layer for spatial weight prediction. The first LSTM takes the sequential flow density representation as input and generates a hidden state at each time-step for attention map inference, while the second LSTM aims at learning the effective spatial-temporal feature expression from attentionally weighted crowd flow features. Based on the ACFM, we further build a deep architecture with the application to citywide crowd flow prediction, which naturally incorporates the sequential and periodic data as well as other external influences. Extensive experiments on two standard benchmarks (i.e., crowd flow in Beijing and New York City) show that the proposed method achieves significant improvements over the state-of-the-art methods.
\end{abstract}

%
%



\copyrightyear{2018} 
\acmYear{2018} 
\setcopyright{acmcopyright}
\acmConference[MM '18]{2018 ACM Multimedia Conference}{October 22--26, 2018}{Seoul, Republic of Korea}
\acmBooktitle{2018 ACM Multimedia Conference (MM '18), October 22--26, 2018, Seoul, Republic of Korea}
\acmPrice{15.00}
\acmDOI{10.1145/3240508.3240636}
\acmISBN{978-1-4503-5665-7/18/10}

\maketitle

\section{Introduction}

Crowd flow prediction is crucial for traffic management and public safety, and has drawn a lot of research interests due to its huge potentials in many intelligent applications, including intelligent traffic diversion and travel optimization.

\begin{figure}[t]
  \centerline{
    \includegraphics[width=0.600\columnwidth]{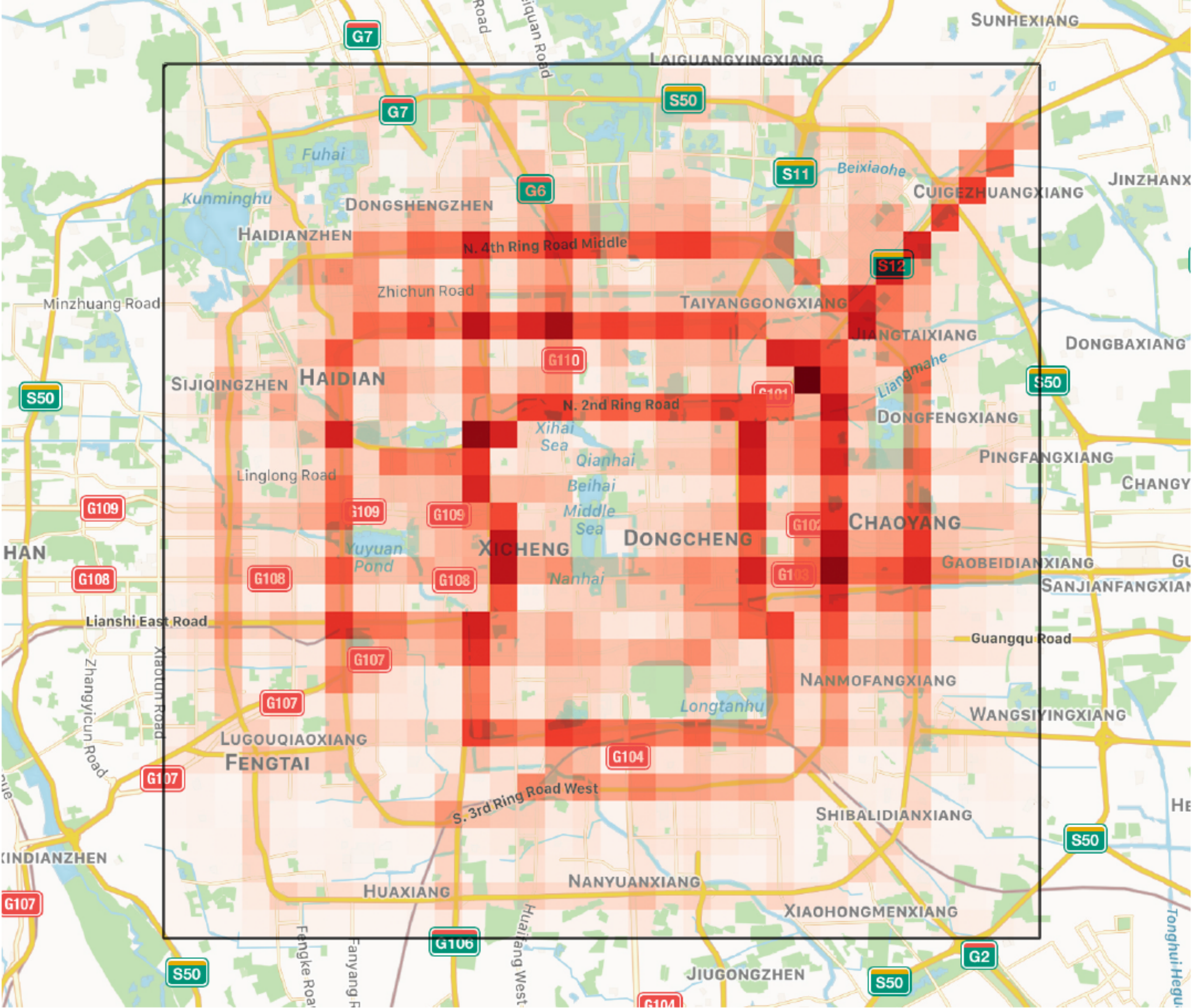}
    \includegraphics[width=0.382\columnwidth]{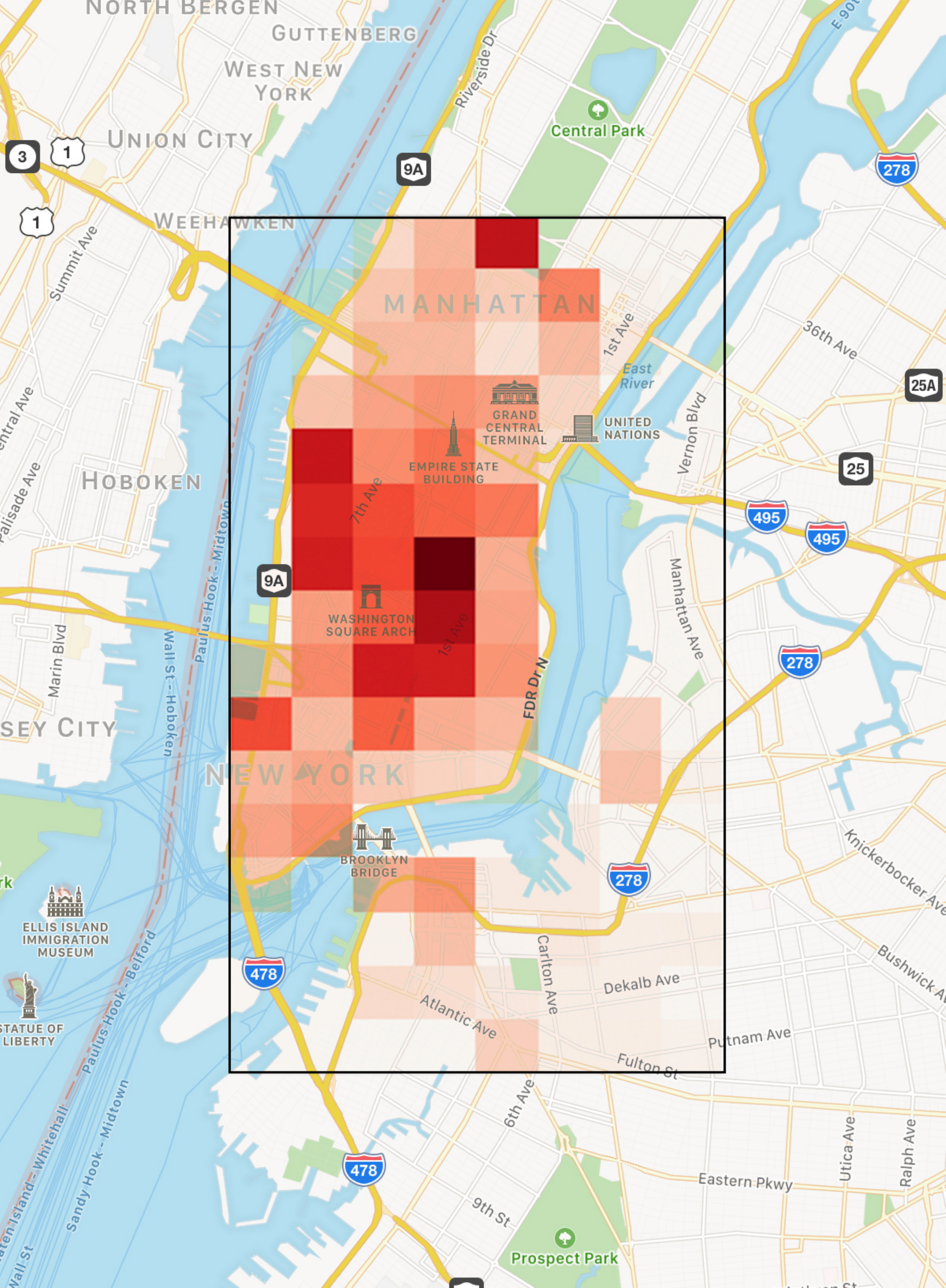}}
  \vspace{-2 mm}
  \caption{Visualization of two crowd flow maps in Beijing and New York City. We partition a city into a grid map based on the longitude and latitude and generate the crowd flow maps by measuring the number of crowd in each region with mobility data (e.g., GPS signals or mobile phone signals). The weight of each grid indicates the flow density of a time period at a specific area.}
  \vspace{-1mm}
\label{fig:citywide_crow_flow_visual}
\end{figure}

Nowadays, we live in an era where ubiquitous digital devices are able to broadcast rich information about human mobility in real-time and at a high rate, which exponentially increases the availability of large-scale mobility data (e.g., GPS signals or mobile phone signals). In this paper, we generate the crowd flow maps from these mobility data and utilize the historical crowd flow maps to forecast the future crowd flow of a city. As shown in Fig.~\ref{fig:citywide_crow_flow_visual}, we partition a city into a grid map based on the longitude and latitude, and measure the number of pedestrians in each region at each time interval with the mobility data. Although the regional scale can vary greatly in different cities, the core problem lies in excavating the evolution of traffic flow in different spatial and temporal regions.

Recently, notable successes have been achieved for citywide crowd flow prediction based on deep neural networks coupled with certain spatial-temporal priors~\cite{zhang2016dnn,zhang2017deep}. Nevertheless, there still exist several challenges limiting the performance of crowd flow analysis in complex scenarios. First, crowd flow data can vary greatly in temporal sequence and capturing such dynamic variations is non-trivial. Second, some periodic laws (e.g., traffic flow suddenly changing due to rush hours or pre-holiday effects) can greatly affect the situation of crowd flow, which increases the difficulty in learning crowd flow representations from data.

To solve all above issues, we propose a novel spatial-temporal neural network module, called Attentive Crowd Flow Machine~(ACFM), to adaptively exploit diverse factors that affect crowd flow evolution and at the same time produce the crowd flow estimation map end-to-end in a unified module. The attention mechanism embedded in ACFM is designed to automatically discover the regions with primary positive impacts for the future flow prediction and simultaneously adjust the impacts of the different regions with different weights at each time-step. Specifically, the ACFM comprises two progressive ConvLSTM~\cite{xingjian2015convolutional} units. The first one takes input from i) the feature map representing flow density at each moment and ii) the memorized representations of previous moments, to compute the attentional weights, while the second LSTM aims at generating superior spatial-temporal feature representation from attentionally weighted sequential flow density features.

The proposed ACFM has the following appealing properties. First, it can effectively incorporate spatial-temporal information in feature representation and can flexibly compose solutions for crowd flow prediction with different types of input data. Second, by integrating the deep attention mechanism~\cite{sharma2015action,lu2016knowing}, ACFM adaptively learns to represent the weights of each spatial location at each time-step, which allows the model to dynamically perceive the impact of the given area at a given moment for the future traffic flow. Third, ACFM is a general and differentiable module which can be effectively combined with various network architectures for end-to-end training.

In addition, for forecasting the citywide crowd flow, we further build a deep architecture based on the ACFM, which consists of three components: i) sequential representation learning, ii) periodic representation learning and iii) a temporally-varying fusion module. The first two components are implemented by two parallel ACFMs for contextual dependencies modeling at different temporal scales, while the temporally-varying fusion module is proposed to adaptively merge the two separate temporal representation for crowd flow predictions.

The main contributions of this work are three-fold:
\begin{itemize}
\item We propose a novel ACFM neural network, which incorporates two LSTM modules with spatial-temporal attentional weights, to enhance the crowd flow prediction via adaptively weighted spatial-temporal feature modeling.
\item We integrate ACFM in our customized deep architecture for citywide crowd flow estimation, which recurrently incorporates various sequential and periodic dependencies with temporally-varying data.
\item Extensive experiments on two public benchmarks of crowd flow prediction demonstrate that our approach outperforms existing state-of-the-art methods by large margins.
\end{itemize}

\section{Related Work}
\textbf{Crowd Flow Analysis}.
Due to the wide application of traffic congestion analysis and public safety monitoring, citywide crowd flow analysis has recently attracted a wide range of research interest~\cite{zheng2014urban}. A pioneer work was proposed by Zheng et al.,~\cite{zheng2015trajectory}, in which they proposed to represent public traffic trajectories as graphs or tensor structures.
Inspired by the significant progress of deep learning on various tasks~\cite{zhang2015bit,chen2016disc,li2017face,zhang2018hierarchical,liu2018facial}, many researchers also have attempted to handle this task with deep neural network. Fouladgar et al.~\cite{fouladgar2017scalable} introduced a scalable decentralized deep neural networks for urban short-term traffic congestion prediction. In~\cite{zhang2016dnn}, a deep learning based framework was proposed to leverage the temporal information of various scales (i.e. temporal closeness, period and seasonal) for crowd flow prediction. Following this work, Zhang et al.,~\cite{zhang2017deep} further employed a convolution based residual network to collectively predict inflow and outflow of crowds in every region of a city grid-map. To take more efficient temporal modeling into consideration, Dai et al.~\cite{dai2017deeptrend} proposed a deep hierarchical neural network for traffic flow prediction, which consists of an extraction layer to extract time-variant trend in traffic flow and a prediction layer for final crowd flow forecasting. Currently, to overcome the scarcity of crowd flow data, Wang et al.~\cite{wang2018crowd} proposed to learn the target city model from the source city model with a region based cross-city deep transfer learning algorithm.

\textbf{Memory and attention neural networks}.
Recurrent neural networks~(RNN) have been widely applied to various sequential prediction tasks~\cite{sutskever2014sequence,donahue2015long}. As a variation of RNN, Long Short-Term Memory Networks~(LSTM) enables RNNs to store information over extended time intervals and exploit longer-term temporal dependencies. It was first applied to the research field of natural language processing~\cite{luong2015effective} and speech recognition~\cite{graves2013speech}, while recently many researchers have attempted to combine CNN with LSTM to model the spatial-temporal information for various of computer vision applications, such as video salient object detection~\cite{li2018flow}, image caption~\cite{mao2014deep,wu2018interpretable} and action recognition~\cite{veeriah2015differential}. Visual attention is a fundamental aspect of human visual system, which refers to the process by which humans focus the computational resources of their brain's visual system to specific regions of the visual field while perceiving the surrounding world. It has been recently embedded in deep convolution networks~\cite{chen2016attention} or recurrent neural networks to adaptively attend on mission-related regions while processing feedforward operation and have been proved effective for many tasks, including machine translation~\cite{luong2015effective}, crowd counting~\cite{liu2018crowd}, multi-label image classification~\cite{wang2017multi}, face hallucination~\cite{cao2017attention}, and visual question answering~\cite{xu2016ask}. However, no existing work incorporates attention mechanism in crowd flow prediction.

\begin{figure*}[t]
  \begin{center}
     \includegraphics[width=1.99\columnwidth]{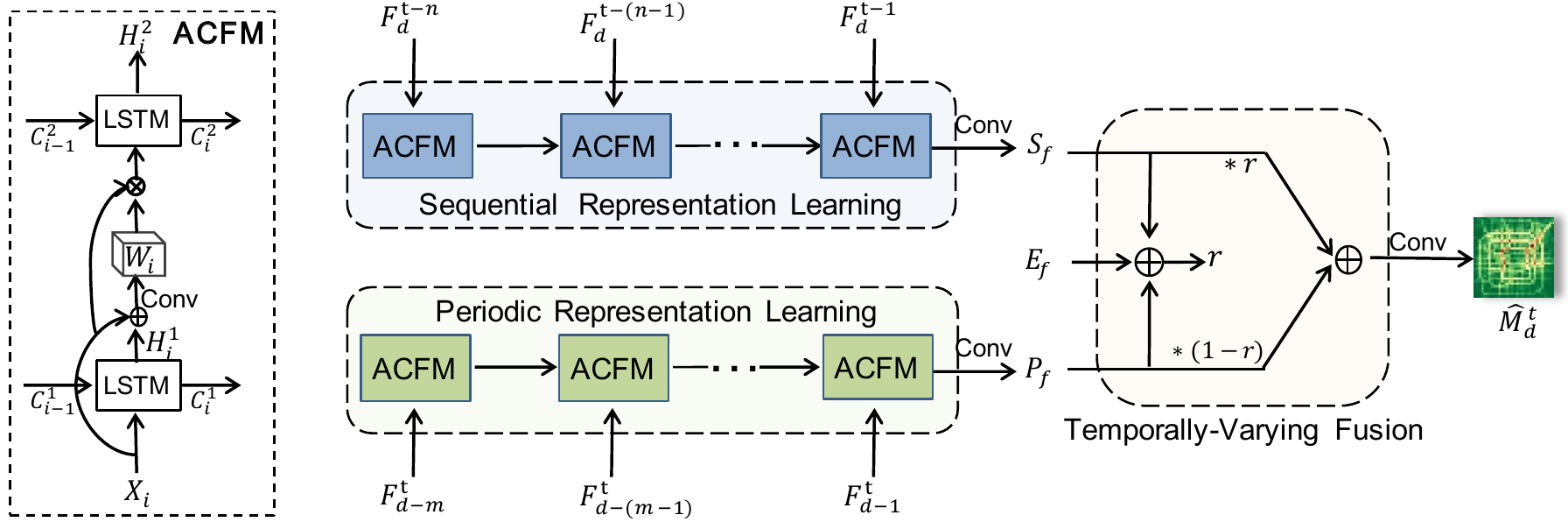}
  \vspace{-4mm}
  \end{center}
   \caption{\textbf{Left:} The architecture of the proposed Attentive Crowd Flow Machine~(ACFM). ACFM can be applied to adequately capture various contextual dependencies for crowd flow evolution analysis. ${X_i}$ denotes the input feature map of the ${i^{th}}$ iteration. ``$\bigoplus$'' denotes feature concatenation and ``$\bigotimes$'' refers to element-wise multiplication.
   \textbf{Right:} The architecture of the our citywide crowd flow prediction networks. It consists of sequential representation learning, periodic representation learning and a temporally-varying fusion module. ${F^t_d}$ denotes the embedding feature of crowd flow and external factors at the ${t^{th}}$ time interval of the ${d^{th}}$ day. ${\widehat{M}^t_d}$ is the predicted crowd flow map. ${S_{f}}$ and ${P_{f}}$ are sequential representation and periodic representation, while external factors integrative feature ${E_{f}}$ is the element-wise addition of external factors features of all relative time intervals. The symbols ${r}$ and ${(1-r)}$ reflect the importance of ${S_{f}}$ and ${P_{f}}$ respectively.}
\vspace{0mm}
\label{fig:network-structure}
\end{figure*}

The most relevant works to us are~\cite{zhang2017fcn,xiong2017spatiotemporal}, which also incorporate ConvLSTM for spatial-temporal modeling. However, they are used for consecutive video frames representation and aims to estimate the crowd counting on a given surveillance image instead of forecasting crowd flow evolution based on mobility data. Moreover, our proposed ACFM is composed of two progressive LSTM modules with learnable attention weights, which is not only adept at modeling spatial-temporal representation, but also efficiently capturing the effect on the global crowd flow evolution caused by the changes of traffic conditions in each particular spatial-temporal region~(e.g. a traffic jam caused by an accident). Last but not the least, the attention mechanism embedded in our ACFM module also helps to improve the interpretability of the network process while boosting the performance.

\section{Preliminaries}
In this section, we first describe some basic elements of crowd flow and then define the crowd flow prediction problem.

\textbf{Region Partition}:
There are many ways to divide a city into multiple regions in terms of different granularities and semantic meanings, such as road network ~\cite{deng2016latent} and zip code tabular. In this study, following the previous works~\cite{zhang2017deep,yao2018deep}, we partition a city into $h \times w$ non-overlapping grid map based on the longitude and latitude. Each rectangular grid represents a different geographical region in the city. Figure~\ref{fig:citywide_crow_flow_visual} illustrates the partitioned regions of Beijing and New York City.

\textbf{Crowd Flow}:
In practical application, we can extract a mass of crowd trajectories from GPS signals or mobile phone signals.  With those crowd trajectories, we can measure the number of pedestrians entering or leaving a given region at each time interval, which are called inflow and outflow in our work. For convenience, we denote the crowd flow map at the ${t^{th}}$ time interval of ${d^{th}}$ day as a tensor ${\bm{M_d^t} \in R^{2\times h\times w}}$, of which the first channel is the inflow and the second channel is the outflow. Some crowd flow maps are visualized in Figure~\ref{fig:predicted_flows_visual}.

\textbf{External Factors}:
As mentioned in the previous work~\cite{zhang2017deep}, crowd flow can be affected by many complex external factors, such as meteorology information and holiday information. In this paper, we also consider the effect of these external factors. The meteorology information (e.g., weather condition, temperature and wind speed) can be collected from some public meteorological websites, such as Wunderground\footnote{\url{https://www.wunderground.com/}}. Specifically, the weather condition is categorized into sixteen categories (e.g., sunny and rainy) and it is digitized with One-Hot Encoding~\cite{harris2010digital}, while temperature and wind speed are scaled into the range [0, 1] with min-max linear normalization. Multiple categories of holiday \footnote{The categories of holiday are variational in different datasets.} (e.g., Chinese Spring Festival and Christmas) can be acquired from calendar and encoded into a binary vector with One-Hot Encoding. Finally, we concatenate all external factors data to a 1D tensor. The external factors tensor at the ${t^{th}}$ time interval of ${d^{th}}$ day is expressed as a ${\bm{E_d^t}}$ in the following sections.

\textbf{Crowd Flow Prediction}:
This problem aims to predict the crowd flow map ${M_d^{t}}$, given historical crowd flow maps and external factors data until the ${(t-1)^{th}}$ time interval of ${d^{th}}$ day.

\section{Attentive Crowd Flow Machine}
We propose a unified neural network module, named Attentive Crowd Flow Machine~(ACFM), to learn the crowd flow spatial-temporal representations. ACFM is designed to adequately capture various contextual dependencies of the crowd flow, e.g., the spatial consistency and the temporal dependency of long and short term. As shown on the left of Fig.~\ref{fig:network-structure}, the ACFM is composed of two progressive ConvLSTM~\cite{xingjian2015convolutional} units connected with a convolutional layer for attention weight prediction at each time step. The first LSTM~(bottom LSTM in the figure) models the temporal dependency through original crowd flow feature embedding~(extracted from CNN), the output hidden state of which is concatenated with current crowd flow feature and fed to a convolution layer for weight map inference. The second LSTM~(upper LSTM in the figure) is of the same structure as the first LSTM but takes the re-weighted crowd flow features as input at each time-step and is trained to recurrently learn the spatial-temporal representations for further crowd flow prediction.

For better understanding, we denote the input feature map of the ${i}^{th}$ iteration as ${X_i \in R^{c\times h\times w}}$, with $h$, $w$ and $c$ representing the height, width and the number of channels. Following ~\cite{hochreiter1997long}, the hidden state ${H_i^1 \in R^{c\times h\times w}}$ of first LSTM can
be formulated as:
 \begin{equation}
\label{equ:first_lstm}
H_i^1 = \textbf{ConvLSTM}(H_{i-1}^1, C_{i-1}^1, X_i),
\end{equation}
where ${C_{i-1}^1}$ is the memorized cell state of the first LSTM at $(i-1)^{th}$ iteration. The internal hidden state ${H_i^1}$ is maintained to model the dynamic temporal behavior of the previous crowd flow sequences.

We concatenate ${H_i^1}$ and ${X_i}$ to generate a new tensor, and feed it to a single convolutional layer with kernel size ${1\times 1}$ to generate an attention map ${W_i}$, which can be expressed as:
\begin{equation}
\label{equ:attention}
W_i = \textbf{Conv}_{1\times 1}(H_i^1 \oplus X_i, w),
\end{equation}
where $\oplus$ denotes feature concatenation and $w$ is the parameters of the convolutional layer. And ${W_i}$ indicates the weights of each spatial location on the feature map ${X_i}$. We further reweigh ${X_i}$ with an element-wise multiplication according to ${W_i}$ and take the reweighed map as input to the second LSTM for representation learning, the hidden state ${H_i^2 \in R^{c\times h\times w}}$ of which can be formulated as:
\begin{equation}
\label{equ:second_lstm}
H_i^2 = \textbf{ConvLSTM}(H_{i-1}^2, C_{i-1}^2, X_i * W_i),
\end{equation}
where ${*}$ refers to the element-wise multiplication. ${h_i^2}$ encodes the attention-aware content of current input as well as memorizes the contextual knowledge of previous moments. The output of the last hidden state thus encodes the information of the whole crowd flow sequence, and is used as the spatial-temporal representation for evolution analysis of future flow map. In the next section, we will show how to incorporate the proposed ACFM in our crowd flow prediction framework.

\section{Citywide Crowd Flow Prediction} ~\label{sec:ccfp}
We build a deep neural network architecture incorporated with our proposed ACFM to predict citywide crowd flow. As illustrated on the right of Fig.~\ref{fig:network-structure}, the crowd flow prediction framework consists of three components: (1) sequential representation learning, (2) periodic representation learning and (3) a temporally-varying fusion module. For the first two parts of the framework, we employ the ACFM to model the contextual dependencies of crowd flow at different temporal scales. After that, a temporally-varying fusion module is proposed to adaptively merge the different feature embeddings from each component with the weight $r$ learned from the concatenation of respective feature representations and the external information. Finally, the merged feature map is fed to one additional convolution layer for crowd flow map inference.

\subsection{Sequential Representation Learning}~\label{sec:srl}
The evolution of citywide crowd flow is usually affected by diverse internal and external factors, e.g., current urban traffic and weather conditions. For instance, a traffic accident occurring on a city main road at 9 am may seriously affect the crowd flow of nearby regions in subsequent time periods. Similarly, a sudden rain may seriously affect the crowd flow in a specific region. To deal with these issues, we take several continuous crowd flow features and their corresponding external factors features as the sequential temporal features, and feed them into our ACFM to recurrently capture the trend of crowd flow in the short term.

Specifically, we denote the input sequential temporal features as:
\begin{equation}
S_{in} = \{F^{t-k}_d \big| k = n, n-1,...,1\},
\end{equation}
where ${n}$ is the length of the sequentially related time intervals and ${F_j^i}$ denotes the embedding features of the crowd flow and the external factors at the $i^{th}$ time interval of the ${j^{th}}$ day. The extraction of embedding feature ${F_j^i}$ will be described in Section~\ref{sec:impl_detail}.
We apply the proposed ACFM to learn sequential representation from temporal features $S_{in}$. As shown on the right of Fig.~\ref{fig:network-structure}, the ACFM recurrently takes each element of $S_{in}$ as input and learns to selectively memorize the context of this specific temporally-varying data.
The output hidden state of the last iteration is further fed into a following convolution layer to generate a feature representation of size ${c\times h\times w}$, denoted as ${S_{f}}$, which forms the spatial-temporal feature embedding of the fine-grained sequential data.

\subsection{Periodic Representation Learning}
Generally, there exist some periodicities which make a significant impact on the changes of traffic flow. For example, the traffic conditions are very similar during morning rush hours of consecutive workdays, repeating every 24 hours. Similar with sequential representation learning described in Section~\ref{sec:srl}, we take periodic temporal features
\begin{equation}
P_{in} = \{F^{t}_{d-k} \big| k = m, m-1,...,1\},
\end{equation}
to capture the periodic property of crowd flow, where ${n}$ is the length of the periodic days. As shown on the right of Fig.~\ref{fig:network-structure}, we employ ACFM to learn periodic representation with the periodic temporal features ${P}$ as input. The hidden output of the last iteration of ACFM is passed through a convolutional layer to generate a representation ${P_{f} \in R^{c\times h\times w} }$. The ${P_{f}}$ encodes the context of periodic laws, which is essential for crowd flow prediction.

\subsection{Temporally-Varying Fusion} ~\label{sec:fusion}
The future crowd flow is affected by the two temporally-varying representations ${S_{f}}$ and ${P_{f}}$. A naive method is to directly merge those two representations, however it is suboptimal. In this subsection, we propose a novel temporally-varying fusion module to adaptively fuse the sequential representation ${S_{f}}$ and the periodic representation ${P_{f}}$ of crowd flow with different weight.

Considering that the external factors may affect the importance proportion of two representations, we take the sequential representation ${S_{f}}$, periodic representation ${P_{f}}$ and the external factors integrative feature ${E_{f}}$ to calculate the fusion weight, where ${E_{f}}$ is the element-wise addition of external factors features of all relative time intervals and will be described in Section~\ref{sec:impl_detail}. As shown on the right of Fig.~\ref{fig:network-structure}, we first concatenate ${S_{f}}$, ${P_{f}}$ and ${E_{f}}$ and feed them as input to two fully-connected layers~(the first layer has 512 neurons and the second has only one neuron) for fusion weight inference. After a sigmoid function, the temporally-varying fusion module outputs a single value $r \in [0,1]$, which reflects the importance of the sequential representation ${S_{f}}$. And $1-r$ is treated as the fusion weight of periodic representation ${P_{f}}$.

We then merge these two temporal representations with different weight and further reduce the feature to two channels~(input and output flow) with a linear transformation, which can be expressed as:
\begin{equation}
\label{equ:fusion}
M_f= \mathcal T(r*S_{f} + (1-r)*P_{f}).
\end{equation}
where ${\mathcal T}$ is the linear transformation implemented by a convolution layer with two filters. The predicted crowd flow map ${\widehat{M}^t_d \in R^{2\times h\times w}}$ can be computed as
\begin{equation}
\label{equ:forecast}
\widehat{M}^t_d = tanh(M_f).
\end{equation}
where the hyperbolic tangent ${tan}$ ensures the output values are within $[-1, 1]$ \footnote{When training, we use Min-Max linear normalization method to scale the crowd flow maps into the range $[-1,1]$. When evaluating, we re-scale the predicted value back to the normal values and then compare with the ground truth.}.

\subsection{Implementation Details} ~\label{sec:impl_detail}
We first detail the method of extracting crowd flow feature as well as external factors feature and then describe our network optimization.

\textbf{Crowd Flow Feature}:
For the crowd flow map ${M^i_j}$ at the ${i^{th}}$ time interval of the ${j^{th}}$ day, we extract its feature ${F^i_j(M)}$ with a customized
ResNet~\cite{he2016deep} structure, which is stacked by ${N}$ residual units without any down-sampling operations. Each residual unit contains two convolutional layers followed by two ReLu layers. We set the channel numbers of all convolutional layers as 16 and the kernel sizes as ${3\times3}$.

\textbf{External Factors Feature}:
For the external factors ${E^i_j}$, we extract its feature with a simple neural network implemented by two fully-connected layers. The first FC layer has 256 neurons and the second one has ${16 \times  h \times  w}$ neurons. The output of the last layer is further reshaped to a 3D tensor ${F^i_j(E) \in R^{16\times h\times w}}$, which is the final feature of ${E^i_j}$.

Finally, we concatenate ${F^i_j(M)}$ and ${F^i_j(E)}$ to generate the embedding feature ${F_j^i}$, which can be expressed as
\begin{equation}
F_j^i= F^i_j(M) \oplus F^i_j(E),
\end{equation}
where $\oplus$ denotes feature concatenation. For the external factors integrative feature ${E_{f}}$ described in Section~\ref{sec:fusion}
, it is the element-wise addition of ${\{E^{t-k}_{d} \big| k = n, n-1,...,1\}}$ and ${\{E^{t}_{d-k} \big| k = m, m-1,...,1\}}$.

\textbf{Network Optimization}:
We adopt the TensorFlow~\cite{abadi2016tensorflow} toolbox to implement our crowd flow prediction network.
The filter weights of all convolutional layers and fully-connected layers are initialized by Xavier~\cite{glorot2010understanding}.
The size of a minibatch is set to 64 and the learning rate is ${10^{-4}}$. We optimize our networks parameters in an end-to-end manner via Adam optimization~\cite{kingma2014adam} by minimizing the Euclidean loss for 270 epochs with a GTX 1080Ti GPU.

\section{Experiments}
In this section, we first conduct experiments on two public benchmarks (e.g., TaxiBJ~\cite{zhang2016dnn} and BikeNYC~\cite{zhang2016dnn}) to evaluate the performance of our model on citywide crowd flow prediction. We further conduct an ablation study to demonstrate the effectiveness of each component in our model.

\subsection{Dataset Setting and Evaluation Metric}
We forecast the inflow and outflow of citywide crowds on two datasets: the TaxiBJ~\cite{zhang2016dnn} dataset for taxicab flow prediction and the BikeNYC~\cite{zhang2016dnn} dataset for bike flow prediction.

{\textbf{TaxiBJ Dataset:}} This dataset contains 22,459 time intervals of crowd flow maps with a size of ${2\times32\times32}$, which are generated with Beijing taxicab GPS trajectory data. The external factors contain weather conditions, temperature, wind speed and 41 categories of holiday. For the fair comparison, we refer to~\cite{zhang2017deep} and take the data in the last four weeks as the testing set and the rest as the training set. In this dataset, we set the sequential length ${\emph n}$ and the periodic length ${\emph m}$ as $3$ and $2$, respectively. As with ST-ResNet~\cite{zhang2017deep}, the ResNet described in Section~\ref{sec:impl_detail} is composed of 12 residual units.

{\textbf{BikeNYC Dataset:}} This dataset is generated with the NYC bike trajectory data, which contains 4,392 available time intervals crowd flow maps with the size of ${2\times16\times8}$. The data of the last ten days are chosen to be the test set. As for external factors, 20 categories of the holiday are recorded. In this dataset, we set the sequential length ${\emph n}$ as 5 and the periodic length ${\emph m}$ as 7. For a fair comparison with ST-ResNet~\cite{zhang2017deep}, we also utilize a ResNet described in Section~\ref{sec:impl_detail} with 4 residual units to extract the crowd flow feature.

We adopt Root Mean Square Error (RMSE) as evaluation metric to evaluate the performances of all the methods, which is defined as:
\begin{small}
 \begin{equation}
RMSE = \sqrt{{\frac{1}{z}}{\sum_{i=1}^z{(\widehat{Y_i} - Y_i)}^2}},
\end{equation}
\end{small}
where ${\widehat{Y_i}}$ and ${Y_i}$ represent the predicted flow map and its ground truth map, respectively. $z$ indicates the number of samples used for validation.

\subsection{Comparison with the State of the Art}

\begin{table}[t]
\newcommand{\tabincell}[2]{\begin{tabular}{@{}#1@{}}#2\end{tabular}}
  \centering
    \begin{tabular}{c|c|c}
    \hline
    \tabincell{c}{Model} & \tabincell{c}{TaxiBJ} & \tabincell{c}{BikeNYC} \\
    \hline\hline
    SARIMA~\cite{williams1998urban} & 26.88 & 10.56 \\
    \hline
    VAR~\cite{lutkepohl2011vector}    & 22.88& 9.92 \\
    \hline
    ARIMA~\cite{box2015time}  & 22.78 & 10.07 \\
    \hline
    ST-ANN~\cite{zhang2017deep} & 19.57 & - \\
    \hline
    DeepST~\cite{zhang2016dnn} & 18.18 & 7.43 \\
    \hline
    ST-ResNet~\cite{zhang2017deep} & 16.69 & 6.33 \\
    \hline
    Ours & \textbf{15.40} & \textbf{5.64} \\
    \hline
    \end{tabular}
  \vspace{1mm}
    \caption{Quantitative comparisons on TaxiBJ and BikeNYC using RMSE (smaller is better). Our proposed method outperforms the existing state-of-the-art methods on both datasets with a margin.
    }
  \vspace{-5mm}
  \label{tab:BJ_NYC_result}
\end{table}

We compare our method with six state-of-the-art methods, including Auto-Regressive Integrated Moving Average (ARIMA)~\cite{box2015time}, Seasonal ARIMA (SARIMA)~\cite{williams1998urban}, Vector Auto-Regressive (VAR)~\cite{lutkepohl2011vector}, ST-ANN~\cite{zhang2017deep}, DeepST~\cite{zhang2016dnn} and ST-ResNet~\cite{zhang2017deep}. For these compared methods, we use the performances provided by Zhang et al.~\cite{zhang2017deep} as their results.

Table~\ref{tab:BJ_NYC_result} summarizes the performance of the proposed method and other six methods. On TaxiBJ dataset, our method decreases the RMSE from 16.69 to 15.40 when compared with current best model, and achieves a relative improvement of 7.7\%. Our method also boosts the prediction accuracy on BikeNYC,~i.e.,~decreases RMSE from 6.33 to 5.64.
Note that some compared methods, e.g., ST-ANN, DeepST and ST-ResNet, also employ deep learning techniques. Experimental results demonstrate that our proposed ACFM is able to explicitly model the spatial-temporal feature as well as the attention weighting of each spatial influence, which greatly outperforms the state-of-the-art. Some crowd flow prediction maps of our full model on TaxiBJ dataset are shown on the second row of the Fig.~\ref{fig:predicted_flows_visual}. As can be seen, our generated crowd flow map is consistently closest to those of the ground-truth, which is accord with the quantitative RMSE comparison.

\subsection{Ablation Study}~\label{sec:abl_study}
Our full model for citywide crowd flow prediction consists of three components: sequential representation learning, periodic representation learning and temporally-varying fusion module. For convenience, we denote our full model as \textbf{S}equential-\textbf{P}eriodic \textbf{N}etwork (\textbf{SPN}) in the following experiments. To show the effectiveness of each component, we implement seven variants of our full model on the TaxiBJ dataset:
\begin{itemize}
\item \textbf{PCNN:} directly concatenates the periodic features ${P_{in}}$ and feeds them to a convolutional layer with two filters followed by ${tanh}$ to predict future crowd flow;
\item \textbf{SCNN:} directly concatenates the sequential features ${S_{in}}$ and feeds them to a convolutional layer followed by ${tanh}$ to predict future crowd flow;
\item \textbf{PRNN-w/o-Attention:} takes periodic features ${P_{in}}$ as input and learns periodic representation with a LSTM layer to predict future crowd flow;
\item \textbf{PRNN:} takes periodic features ${P_{in}}$ as input and learns periodic representation with the proposed ACFM to predict future crowd flow;
\item \textbf{SRNN-w/o-Attention:} takes sequential features ${S_{in}}$ as input and learns sequential representation with a LSTM layer for crowd flow estimation;
\item \textbf{SRNN:} takes sequential features ${S_{in}}$ as input and learns sequential representation with the proposed ACFM to predict future crowd flow;
\item \textbf{SPN-w/o-Fusion:} directly merges sequential representation and periodic representation with equal weight (0.5) to predict future crowd flow.
\end{itemize}

\begin{table}[t]
\newcommand{\tabincell}[2]{\begin{tabular}{@{}#1@{}}#2\end{tabular}}
  \centering
    \begin{tabular}{c|c}
    \hline
    \tabincell{c}{Model} & \tabincell{c}{RMSE} \\
    \hline\hline
    PCNN & 33.44 \\
    PRNN-w/o-Attention & 32.97  \\
    PRNN & 32.52\\
    \hline
    SCNN & 17.48 \\
    SRNN-w/o-Attention & 16.62  \\
    SRNN & 16.11 \\
    \hline
    SPN-w/o-Fusion & 16.01 \\
    SPN & 15.40  \\
    \hline
    \end{tabular}

  \vspace{1mm}
    \caption{Quantitative comparisons (RMSE) of different variants of our model on TaxiBJ dataset for component analysis.}
  \vspace{-5mm}
  \label{tab:component_results}
\end{table}

\begin{figure*}
\begin{center}
\begin{minipage}[a]{0.85\textwidth}
\includegraphics[width=0.49\textwidth]{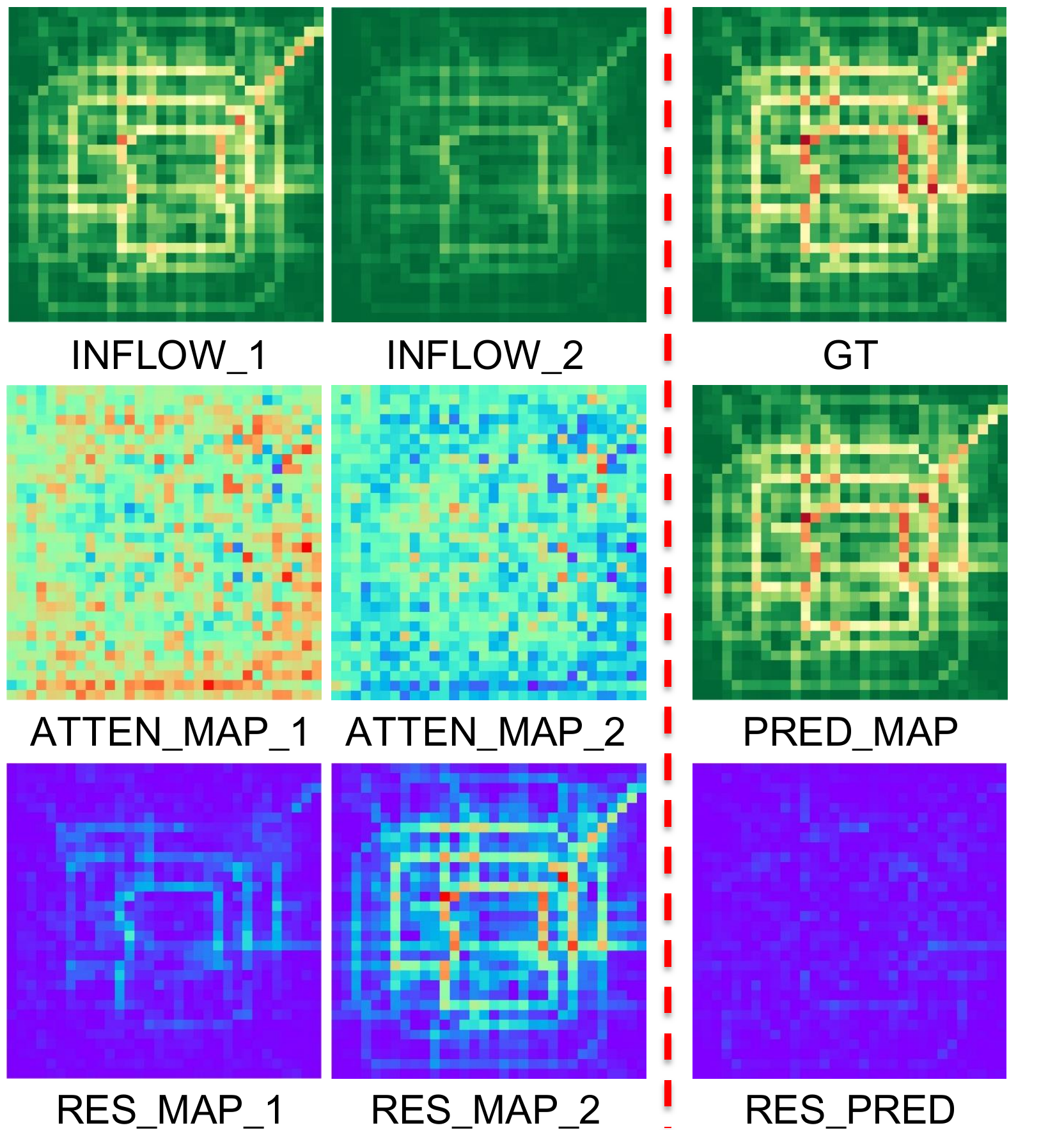}
\includegraphics[width=0.49\textwidth]{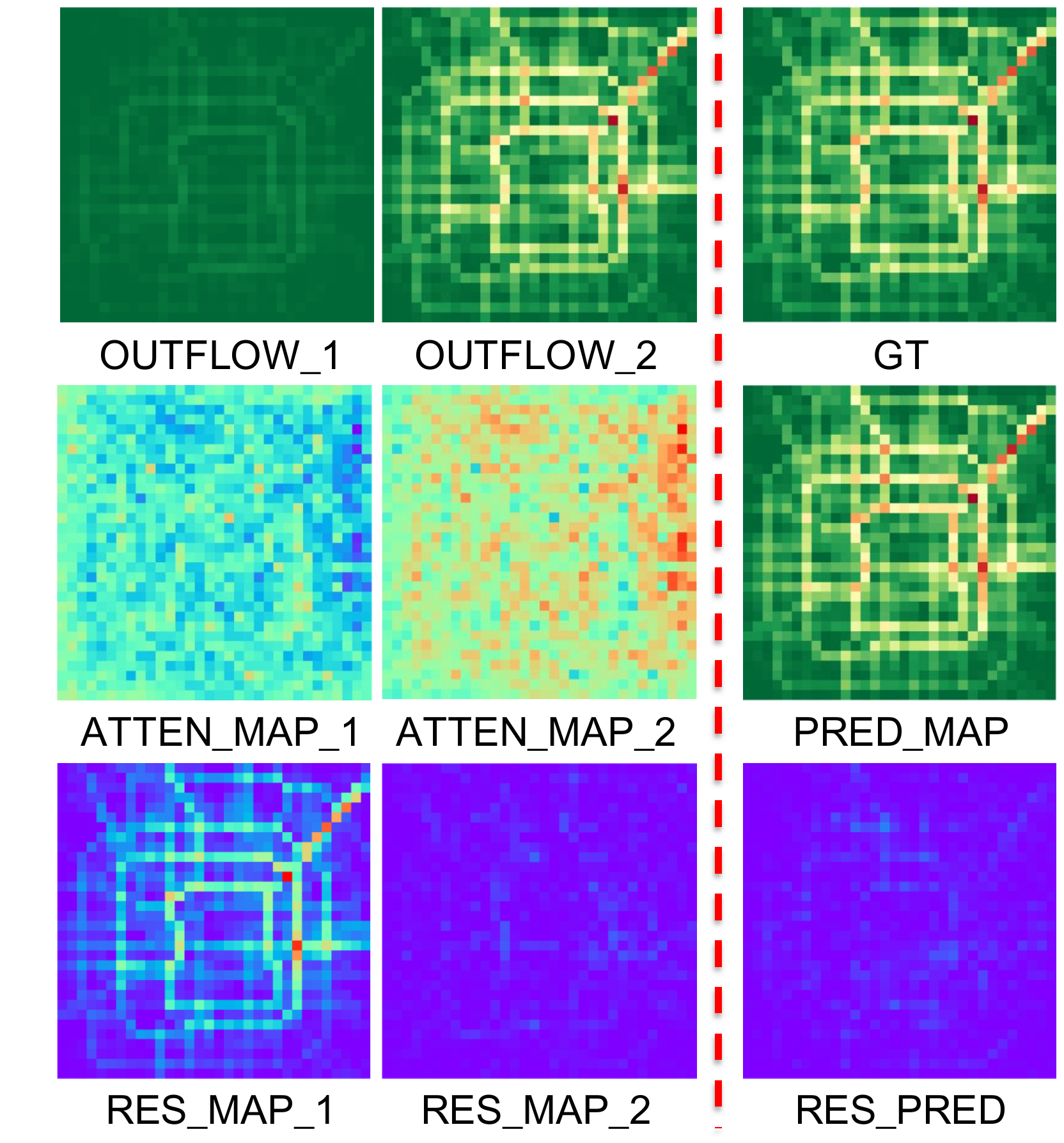}
\end{minipage}
 \vspace{-4mm}
\end{center}
   \caption{Illustration of the generated attentional maps of the crowd flow in {\color{red}{periodic representation learning}} with ${m}$ set as 2. Every three columns form one group. In each group:
   i) on the first row, the first two images are the input periodic inflow/outflow maps and the last one is the ground truth inflow/outflow map of next time interval;
   ii) on the second row, the first two images are the attentional maps generated by our ACFM, while the last one is our predicted inflow/outflow map;
   iii) on the third row, the first two images are the residual maps between the input flow maps and the ground truth, while the last one is the residual map between our predicted flow map and the ground truth.}
\vspace{-0mm}
\label{fig:Periodic-Attention}
\end{figure*}

\begin{figure*}
\begin{center}
\begin{minipage}[a]{1\textwidth}
\includegraphics[width=0.49\textwidth]{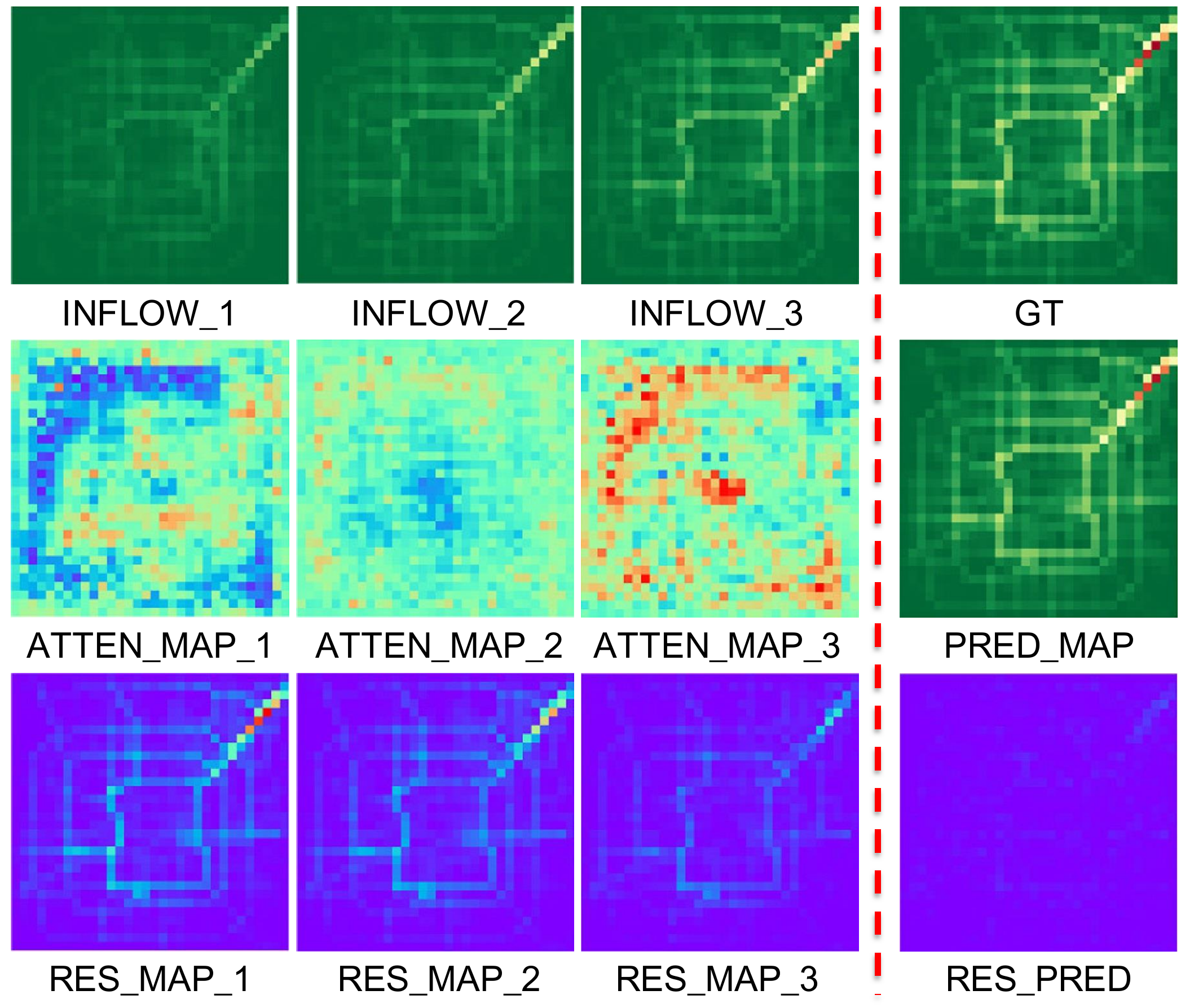}
\includegraphics[width=0.49\textwidth]{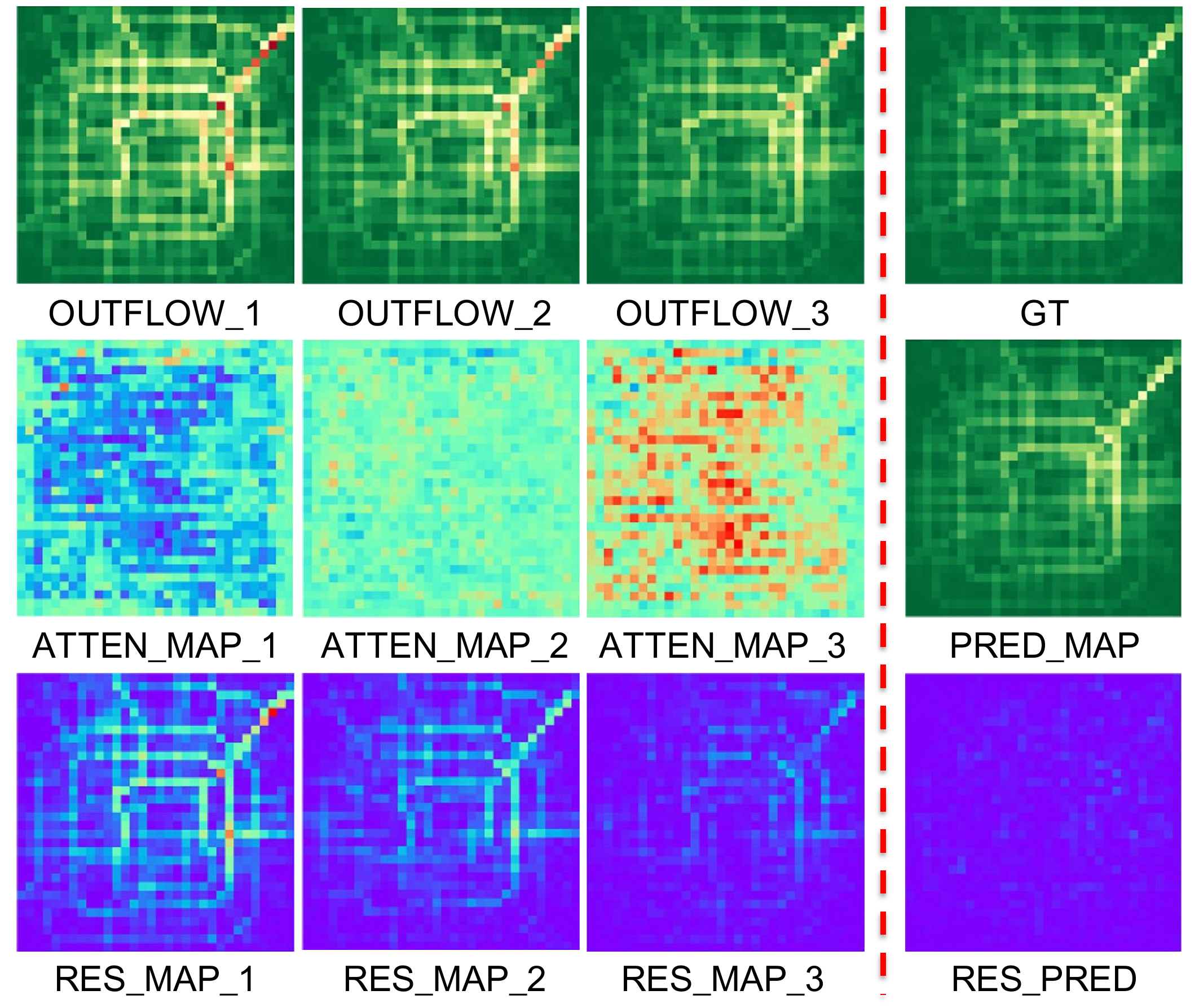}
\end{minipage}
 \vspace{-4mm}
\end{center}
   \caption{Illustration of the generated attentional maps of the crowd flow in {\color{red}{sequential representation learning}} with ${n}$ set as 3. Every four columns form one group. In each group:
   i) on the first row, the first three images are the input sequential inflow/outflow maps and the last one is the ground truth inflow/outflow map of next time interval;
   ii) on the second row, the first three images are the attentional maps generated by our ACFM, while the last one is our predicted inflow/outflow map;
   iii) on the third row, the first three images are the residual maps between the input flow maps and the ground truth, while the last one is the residual map between our predicted flow map and the ground truth.}
\vspace{-0mm}
\label{fig:Sequential-Attention}
\end{figure*}

\textbf{Effectiveness of Spatial Attention:}
As shown in Table~\ref{tab:component_results}, adopting spatial attention, SRNN decreases the RMSE by 0.51, compared to SRNN-w/o-Attention.
For another pair of variants, PRNN with spatial attention has the similar performance improvement, compared to PRNN-w/o-Attention.
Fig.~\ref{fig:Periodic-Attention} and Fig.~\ref{fig:Sequential-Attention} show some attentional maps generated by our method as well as the residual maps between the input crowd flow maps and their corresponding ground truth. We can observe that there is a negative correlation between the attentional maps and the residual maps. It indicates that our ACFM is able to capture valuable regions at each time step and make better predictions by inferring the trend of evolution. Roughly, the greater difference a region has, the smaller its weight, and vice versa. We can inhibit the impacts of the regions with great differences by multiplying the small weights on their corresponding location features. With the visualization of attentional maps, we can also get to know which regions have the primary positive impacts for the future flow prediction. According to the experiment, we can see that the proposed model can not only effectively improve the prediction accuracy, but also enhance the interpretability of the model to a certain extent.

\textbf{Effectiveness of Sequential Representation Learning:}
As shown in Table~\ref{tab:component_results}, directly concatenating the sequential features ${S}$ for prediction, the baseline variant SCNN gets an RMSE of 17.48. When explicitly modeling the sequential contextual dependencies of crowd flow using the proposed ACFM, the variant SRNN decreases the RMSE to 16.11, with 7.8\% relative performance improvement compared to the baseline SCNN, which indicates the effectiveness of the sequential representation learning.

\begin{figure}
\centering
   \includegraphics[width=0.9\columnwidth]{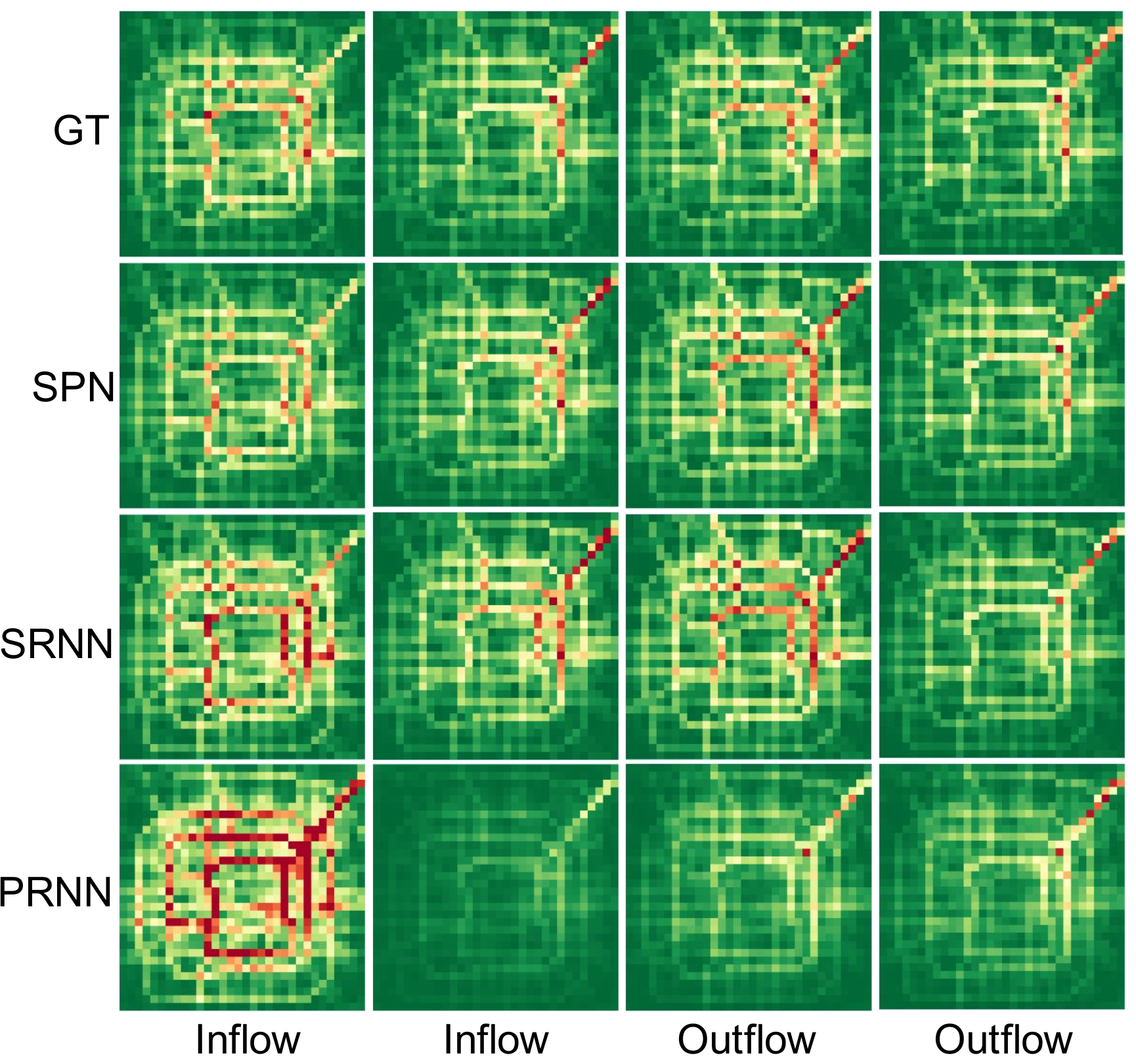}
\vspace{-3mm}
   \caption{Visual comparison of predicted flow maps of different variants on TaxiBJ dataset. The first two columns are inflow maps and the other two columns are outflow maps. The first row is the ground truth maps of crowd flow, while the bottom three rows are the predicted flow maps of SPN, SRNN and PRNN respectively. We can observe that i) the combinations of PRNN and SRNN can help to generate more precise crowd flow maps and ii) the difference between the predicted flow maps of our full model SPN and the ground truth maps are relatively small.}
\vspace{-1mm}
\label{fig:predicted_flows_visual}
\end{figure}

\textbf{Effectiveness of Periodic Representation Learning:}
We also explore different network architectures to learn the periodic representation. As shown in Table~\ref{tab:component_results}, the PCNN, which learns to estimate the flow map by simply concatenating all of the periodic features ${P}$, only achieves RMSE of 33.44. In contrast, when introducing ACFM to learn the periodic representation, the RMSE drops to 32.52. This further demonstrates the effectiveness of the proposed ACFM for spatial-temporal modeling.

\textbf{Effectiveness of Temporally-Varying Fusion:}
When directly merging the two temporal representations with an equal contribution (0.5), SPN-w/o-fusion achieves a negligible improvement, compared to SRNN. In contrast, after using our proposed fusion strategy, the full model SPN decreases the RMSE from 16.11 to 15.40, with a relative
improvement of 4.4\% compared with SRNN. The results show that the significance of these two representations are not equal and are influenced by various factors. The proposed fusion strategy is effective to adaptively merge the different temporal representations and further improve the performance of crowd flow prediction.

\begin{figure}
\centering
   \includegraphics[width=0.90\columnwidth]{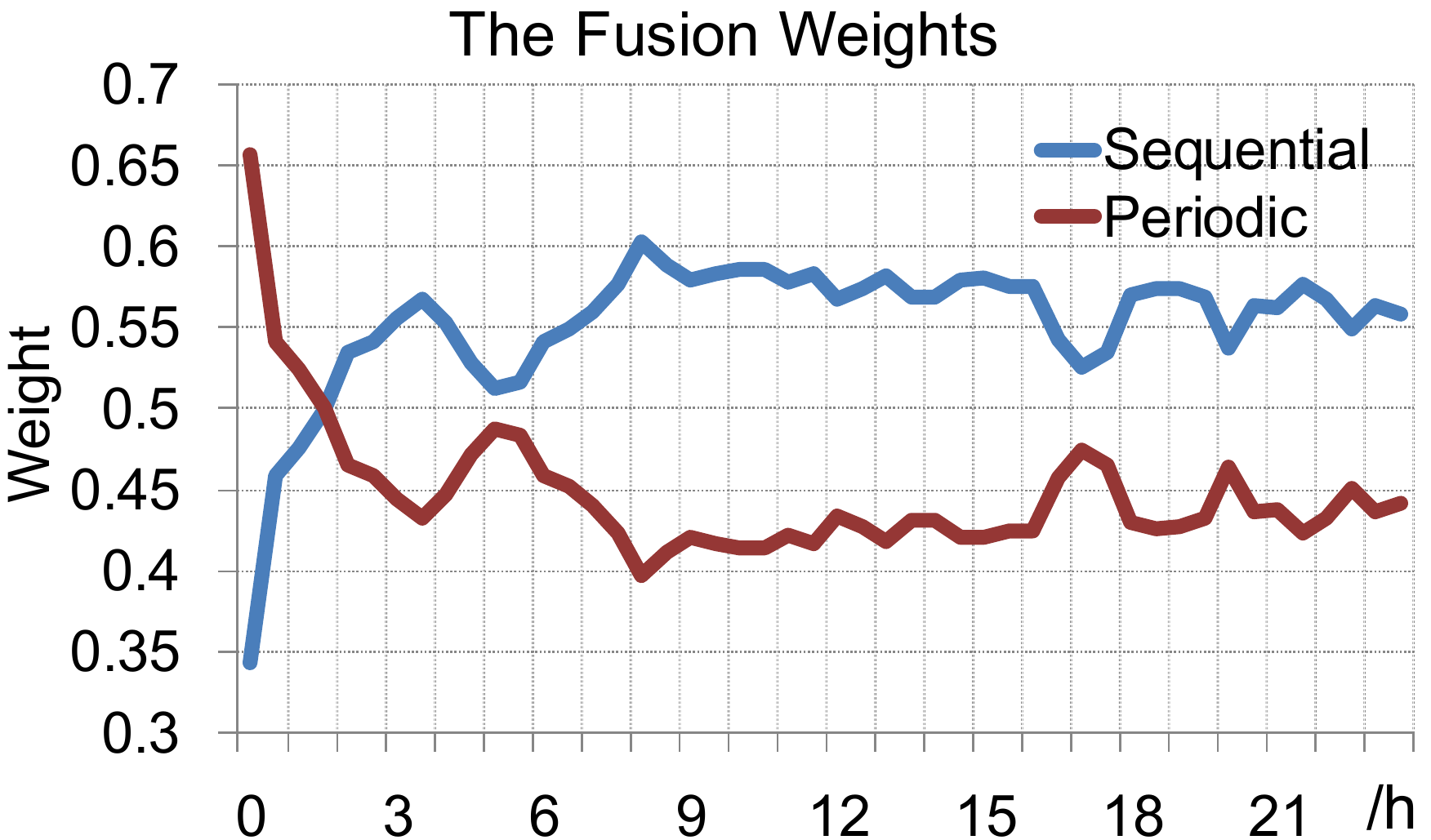}
\vspace{-3mm}
   \caption{The average fusion weights of two types of temporal representation on TaxiBJ testing set.
   We can find that the weights of sequential representation are greater than that of the periodic representation, which indicates the sequential trend is more essential for crowd flow prediction.
   }
\vspace{-3mm}
\label{fig:ratio_rmse_bj}
\end{figure}

\textbf{Further Discussion:}
To analyze how each temporal representation contributes to the performance of crowd flow prediction, we further measure the average fusion weights of two temporal representations at each time interval. As shown in the right of Fig.~\ref{fig:ratio_rmse_bj}, the fusion weights of sequential representation are greater than that of the periodic representation at most time excepting for wee hours. Based on this observation, we can conclude that the sequential representation is more essential for the crowd flow prediction. Although the weight is low, the periodic representation still helps to improve the performance of crowd flow prediction qualitatively and quantitatively.
Fusing with periodic representation, we can decrease the RMSE of SRNN by 4.4\% and generate more precise crowd flow maps, as shown in Table~\ref{tab:component_results} and Fig.~\ref{fig:predicted_flows_visual}.

\section{Conclusion}
This work studies the spatial-temporal modeling for crowd flow prediction problem. To incorporate various factors that affect the flow changes, we propose a unified neural network module named Attentive Crowd Flow Machine~(ACFM). In contrast to the existing flow estimation methods, our ACFM explicitly learns dynamic representations of temporally-varying data with an attention mechanism and can infer the evolution of the future crowd flow from historical crowd flow maps. A unified framework is also proposed to merge two types of temporal information for further prediction. According to the extensive experiments, we have exhaustively verified the effectiveness of our proposed ACFM on the task for citywide crowd flow prediction.


\bibliographystyle{ACM-Reference-Format}
\balance
\bibliography{ACFM-Reference}


\begin{thebibliography}{41}


\ifx \showCODEN    \undefined \def \showCODEN     #1{\unskip}     \fi
\ifx \showDOI      \undefined \def \showDOI       #1{#1}\fi
\ifx \showISBNx    \undefined \def \showISBNx     #1{\unskip}     \fi
\ifx \showISBNxiii \undefined \def \showISBNxiii  #1{\unskip}     \fi
\ifx \showISSN     \undefined \def \showISSN      #1{\unskip}     \fi
\ifx \showLCCN     \undefined \def \showLCCN      #1{\unskip}     \fi
\ifx \shownote     \undefined \def \shownote      #1{#1}          \fi
\ifx \showarticletitle \undefined \def \showarticletitle #1{#1}   \fi
\ifx \showURL      \undefined \def \showURL       {\relax}        \fi
\providecommand\bibfield[2]{#2}
\providecommand\bibinfo[2]{#2}
\providecommand\natexlab[1]{#1}
\providecommand\showeprint[2][]{arXiv:#2}

\bibitem[\protect\citeauthoryear{Abadi, Agarwal, Barham, Brevdo, Chen, Citro,
  Corrado, Davis, Dean, Devin, et~al\mbox{.}}{Abadi et~al\mbox{.}}{2016}]%
        {abadi2016tensorflow}
\bibfield{author}{\bibinfo{person}{Mart{\'\i}n Abadi}, \bibinfo{person}{Ashish
  Agarwal}, \bibinfo{person}{Paul Barham}, \bibinfo{person}{Eugene Brevdo},
  \bibinfo{person}{Zhifeng Chen}, \bibinfo{person}{Craig Citro},
  \bibinfo{person}{Greg~S Corrado}, \bibinfo{person}{Andy Davis},
  \bibinfo{person}{Jeffrey Dean}, \bibinfo{person}{Matthieu Devin},
  {et~al\mbox{.}}} \bibinfo{year}{2016}\natexlab{}.
\newblock \showarticletitle{Tensorflow: Large-scale machine learning on
  heterogeneous distributed systems}.
\newblock \bibinfo{journal}{\emph{arXiv:1603.04467}} (\bibinfo{year}{2016}).
\newblock


\bibitem[\protect\citeauthoryear{Box, Jenkins, Reinsel, and Ljung}{Box
  et~al\mbox{.}}{2015}]%
        {box2015time}
\bibfield{author}{\bibinfo{person}{George~EP Box}, \bibinfo{person}{Gwilym~M
  Jenkins}, \bibinfo{person}{Gregory~C Reinsel}, {and} \bibinfo{person}{Greta~M
  Ljung}.} \bibinfo{year}{2015}\natexlab{}.
\newblock \bibinfo{booktitle}{\emph{Time series analysis: forecasting and
  control}}.
\newblock


\bibitem[\protect\citeauthoryear{Cao, Lin, Shi, Liang, and Li}{Cao
  et~al\mbox{.}}{2017}]%
        {cao2017attention}
\bibfield{author}{\bibinfo{person}{Qingxing Cao}, \bibinfo{person}{Liang Lin},
  \bibinfo{person}{Yukai Shi}, \bibinfo{person}{Xiaodan Liang}, {and}
  \bibinfo{person}{Guanbin Li}.} \bibinfo{year}{2017}\natexlab{}.
\newblock \showarticletitle{Attention-aware face hallucination via deep
  reinforcement learning}.
\newblock \bibinfo{journal}{\emph{arXiv:1708.03132}} (\bibinfo{year}{2017}).
\newblock


\bibitem[\protect\citeauthoryear{Chen, Yang, Wang, Xu, and Yuille}{Chen
  et~al\mbox{.}}{2016b}]%
        {chen2016attention}
\bibfield{author}{\bibinfo{person}{Liang-Chieh Chen}, \bibinfo{person}{Yi
  Yang}, \bibinfo{person}{Jiang Wang}, \bibinfo{person}{Wei Xu}, {and}
  \bibinfo{person}{Alan~L Yuille}.} \bibinfo{year}{2016}\natexlab{b}.
\newblock \showarticletitle{Attention to scale: Scale-aware semantic image
  segmentation}. In \bibinfo{booktitle}{\emph{CVPR}}.
\newblock


\bibitem[\protect\citeauthoryear{Chen, Lin, Liu, Luo, and Li}{Chen
  et~al\mbox{.}}{2016a}]%
        {chen2016disc}
\bibfield{author}{\bibinfo{person}{Tianshui Chen}, \bibinfo{person}{Liang Lin},
  \bibinfo{person}{Lingbo Liu}, \bibinfo{person}{Xiaonan Luo}, {and}
  \bibinfo{person}{Xuelong Li}.} \bibinfo{year}{2016}\natexlab{a}.
\newblock \showarticletitle{DISC: Deep Image Saliency Computing via Progressive
  Representation Learning.}
\newblock \bibinfo{journal}{\emph{TNNLS}} \bibinfo{volume}{27},
  \bibinfo{number}{6} (\bibinfo{year}{2016}), \bibinfo{pages}{1135--1149}.
\newblock


\bibitem[\protect\citeauthoryear{Dai, Fu, Lin, Li, and Wang}{Dai
  et~al\mbox{.}}{2017}]%
        {dai2017deeptrend}
\bibfield{author}{\bibinfo{person}{Xingyuan Dai}, \bibinfo{person}{Rui Fu},
  \bibinfo{person}{Yilun Lin}, \bibinfo{person}{Li Li}, {and}
  \bibinfo{person}{Fei-Yue Wang}.} \bibinfo{year}{2017}\natexlab{}.
\newblock \showarticletitle{DeepTrend: A Deep Hierarchical Neural Network for
  Traffic Flow Prediction}.
\newblock \bibinfo{journal}{\emph{arXiv:1707.03213}} (\bibinfo{year}{2017}).
\newblock


\bibitem[\protect\citeauthoryear{Deng, Shahabi, Demiryurek, Zhu, Yu, and
  Liu}{Deng et~al\mbox{.}}{2016}]%
        {deng2016latent}
\bibfield{author}{\bibinfo{person}{Dingxiong Deng}, \bibinfo{person}{Cyrus
  Shahabi}, \bibinfo{person}{Ugur Demiryurek}, \bibinfo{person}{Linhong Zhu},
  \bibinfo{person}{Rose Yu}, {and} \bibinfo{person}{Yan Liu}.}
  \bibinfo{year}{2016}\natexlab{}.
\newblock \showarticletitle{Latent Space Model for Road Networks to Predict
  Time-Varying Traffic}.
\newblock \bibinfo{journal}{\emph{KDD}} (\bibinfo{year}{2016}),
  \bibinfo{pages}{1525--1534}.
\newblock


\bibitem[\protect\citeauthoryear{Donahue, Anne~Hendricks, Guadarrama, Rohrbach,
  Venugopalan, Saenko, and Darrell}{Donahue et~al\mbox{.}}{2015}]%
        {donahue2015long}
\bibfield{author}{\bibinfo{person}{Jeffrey Donahue}, \bibinfo{person}{Lisa
  Anne~Hendricks}, \bibinfo{person}{Sergio Guadarrama}, \bibinfo{person}{Marcus
  Rohrbach}, \bibinfo{person}{Subhashini Venugopalan}, \bibinfo{person}{Kate
  Saenko}, {and} \bibinfo{person}{Trevor Darrell}.}
  \bibinfo{year}{2015}\natexlab{}.
\newblock \showarticletitle{Long-term recurrent convolutional networks for
  visual recognition and description}. In \bibinfo{booktitle}{\emph{CVPR}}.
\newblock


\bibitem[\protect\citeauthoryear{Fouladgar, Parchami, Elmasri, and
  Ghaderi}{Fouladgar et~al\mbox{.}}{2017}]%
        {fouladgar2017scalable}
\bibfield{author}{\bibinfo{person}{Mohammadhani Fouladgar},
  \bibinfo{person}{Mostafa Parchami}, \bibinfo{person}{Ramez Elmasri}, {and}
  \bibinfo{person}{Amir Ghaderi}.} \bibinfo{year}{2017}\natexlab{}.
\newblock \showarticletitle{Scalable deep traffic flow neural networks for
  urban traffic congestion prediction}.
\newblock \bibinfo{journal}{\emph{arXiv preprint arXiv:1703.01006}}
  (\bibinfo{year}{2017}).
\newblock


\bibitem[\protect\citeauthoryear{Glorot and Bengio}{Glorot and Bengio}{2010}]%
        {glorot2010understanding}
\bibfield{author}{\bibinfo{person}{Xavier Glorot} {and} \bibinfo{person}{Yoshua
  Bengio}.} \bibinfo{year}{2010}\natexlab{}.
\newblock \showarticletitle{Understanding the difficulty of training deep
  feedforward neural networks}. In \bibinfo{booktitle}{\emph{AISTATS}}.
  \bibinfo{pages}{249--256}.
\newblock


\bibitem[\protect\citeauthoryear{Graves, Mohamed, and Hinton}{Graves
  et~al\mbox{.}}{2013}]%
        {graves2013speech}
\bibfield{author}{\bibinfo{person}{Alex Graves}, \bibinfo{person}{Abdel-rahman
  Mohamed}, {and} \bibinfo{person}{Geoffrey Hinton}.}
  \bibinfo{year}{2013}\natexlab{}.
\newblock \showarticletitle{Speech recognition with deep recurrent neural
  networks}. In \bibinfo{booktitle}{\emph{ICASSP}}.
\newblock


\bibitem[\protect\citeauthoryear{Harris and Harris}{Harris and Harris}{2010}]%
        {harris2010digital}
\bibfield{author}{\bibinfo{person}{David Harris} {and} \bibinfo{person}{Sarah
  Harris}.} \bibinfo{year}{2010}\natexlab{}.
\newblock \bibinfo{booktitle}{\emph{Digital design and computer architecture}}.
\newblock \bibinfo{publisher}{Morgan Kaufmann}.
\newblock


\bibitem[\protect\citeauthoryear{He, Zhang, Ren, and Sun}{He
  et~al\mbox{.}}{2016}]%
        {he2016deep}
\bibfield{author}{\bibinfo{person}{Kaiming He}, \bibinfo{person}{Xiangyu
  Zhang}, \bibinfo{person}{Shaoqing Ren}, {and} \bibinfo{person}{Jian Sun}.}
  \bibinfo{year}{2016}\natexlab{}.
\newblock \showarticletitle{Deep residual learning for image recognition}. In
  \bibinfo{booktitle}{\emph{CVPR}}.
\newblock


\bibitem[\protect\citeauthoryear{Hochreiter and Schmidhuber}{Hochreiter and
  Schmidhuber}{1997}]%
        {hochreiter1997long}
\bibfield{author}{\bibinfo{person}{Sepp Hochreiter} {and}
  \bibinfo{person}{J{\"u}rgen Schmidhuber}.} \bibinfo{year}{1997}\natexlab{}.
\newblock \showarticletitle{Long short-term memory}.
\newblock \bibinfo{journal}{\emph{Neural computation}} \bibinfo{volume}{9},
  \bibinfo{number}{8} (\bibinfo{year}{1997}), \bibinfo{pages}{1735--1780}.
\newblock


\bibitem[\protect\citeauthoryear{Kingma and Ba}{Kingma and Ba}{2014}]%
        {kingma2014adam}
\bibfield{author}{\bibinfo{person}{Diederik Kingma} {and}
  \bibinfo{person}{Jimmy Ba}.} \bibinfo{year}{2014}\natexlab{}.
\newblock \showarticletitle{Adam: A method for stochastic optimization}.
\newblock \bibinfo{journal}{\emph{arXiv:1412.6980}} (\bibinfo{year}{2014}).
\newblock


\bibitem[\protect\citeauthoryear{Li, Xie, Wei, Wang, and Lin}{Li
  et~al\mbox{.}}{2018}]%
        {li2018flow}
\bibfield{author}{\bibinfo{person}{Guanbin Li}, \bibinfo{person}{Yuan Xie},
  \bibinfo{person}{Tianhao Wei}, \bibinfo{person}{Keze Wang}, {and}
  \bibinfo{person}{Liang Lin}.} \bibinfo{year}{2018}\natexlab{}.
\newblock \showarticletitle{Flow Guided Recurrent Neural Encoder for Video
  Salient Object Detection}. In \bibinfo{booktitle}{\emph{CVPR}}.
  \bibinfo{pages}{3243--3252}.
\newblock


\bibitem[\protect\citeauthoryear{Li, Liu, Lin, and Wang}{Li
  et~al\mbox{.}}{2017}]%
        {li2017face}
\bibfield{author}{\bibinfo{person}{Ya Li}, \bibinfo{person}{Lingbo Liu},
  \bibinfo{person}{Liang Lin}, {and} \bibinfo{person}{Qing Wang}.}
  \bibinfo{year}{2017}\natexlab{}.
\newblock \showarticletitle{Face Recognition by Coarse-to-Fine Landmark
  Regression with Application to ATM Surveillance}. In
  \bibinfo{booktitle}{\emph{CCCV}}. Springer, \bibinfo{pages}{62--73}.
\newblock


\bibitem[\protect\citeauthoryear{Liu, Li, Xie, Yu, and Lin}{Liu
  et~al\mbox{.}}{2018a}]%
        {liu2018facial}
\bibfield{author}{\bibinfo{person}{Lingbo Liu}, \bibinfo{person}{Guanbin Li},
  \bibinfo{person}{Yuan Xie}, \bibinfo{person}{Yizhou Yu}, {and}
  \bibinfo{person}{Liang Lin}.} \bibinfo{year}{2018}\natexlab{a}.
\newblock \showarticletitle{Facial Landmark Localization in the Wild by
  Backbone-Branches Representation Learning}. In
  \bibinfo{booktitle}{\emph{BigMM}}. IEEE.
\newblock


\bibitem[\protect\citeauthoryear{Liu, Wang, Li, Ouyang, and Lin}{Liu
  et~al\mbox{.}}{2018b}]%
        {liu2018crowd}
\bibfield{author}{\bibinfo{person}{Lingbo Liu}, \bibinfo{person}{Hongjun Wang},
  \bibinfo{person}{Guanbin Li}, \bibinfo{person}{Wanli Ouyang}, {and}
  \bibinfo{person}{Liang Lin}.} \bibinfo{year}{2018}\natexlab{b}.
\newblock \showarticletitle{Crowd Counting using Deep Recurrent Spatial-Aware
  Network}. In \bibinfo{booktitle}{\emph{IJCAI}}.
\newblock


\bibitem[\protect\citeauthoryear{Lu, Xiong, Parikh, and Socher}{Lu
  et~al\mbox{.}}{2016}]%
        {lu2016knowing}
\bibfield{author}{\bibinfo{person}{Jiasen Lu}, \bibinfo{person}{Caiming Xiong},
  \bibinfo{person}{Devi Parikh}, {and} \bibinfo{person}{Richard Socher}.}
  \bibinfo{year}{2016}\natexlab{}.
\newblock \showarticletitle{Knowing when to look: Adaptive attention via A
  visual sentinel for image captioning}.
\newblock \bibinfo{journal}{\emph{arXiv:1612.01887}} (\bibinfo{year}{2016}).
\newblock


\bibitem[\protect\citeauthoryear{Luong, Pham, and Manning}{Luong
  et~al\mbox{.}}{2015}]%
        {luong2015effective}
\bibfield{author}{\bibinfo{person}{Minh-Thang Luong}, \bibinfo{person}{Hieu
  Pham}, {and} \bibinfo{person}{Christopher~D Manning}.}
  \bibinfo{year}{2015}\natexlab{}.
\newblock \showarticletitle{Effective approaches to attention-based neural
  machine translation}.
\newblock \bibinfo{journal}{\emph{arXiv:1508.04025}} (\bibinfo{year}{2015}).
\newblock


\bibitem[\protect\citeauthoryear{L{\"u}tkepohl}{L{\"u}tkepohl}{2011}]%
        {lutkepohl2011vector}
\bibfield{author}{\bibinfo{person}{Helmut L{\"u}tkepohl}.}
  \bibinfo{year}{2011}\natexlab{}.
\newblock \showarticletitle{Vector autoregressive models}.
\newblock In \bibinfo{booktitle}{\emph{International Encyclopedia of
  Statistical Science}}. \bibinfo{pages}{1645--1647}.
\newblock


\bibitem[\protect\citeauthoryear{Mao, Xu, Yang, Wang, Huang, and Yuille}{Mao
  et~al\mbox{.}}{2014}]%
        {mao2014deep}
\bibfield{author}{\bibinfo{person}{Junhua Mao}, \bibinfo{person}{Wei Xu},
  \bibinfo{person}{Yi Yang}, \bibinfo{person}{Jiang Wang},
  \bibinfo{person}{Zhiheng Huang}, {and} \bibinfo{person}{Alan Yuille}.}
  \bibinfo{year}{2014}\natexlab{}.
\newblock \showarticletitle{Deep captioning with multimodal recurrent neural
  networks (m-rnn)}.
\newblock \bibinfo{journal}{\emph{arXiv:1412.6632}} (\bibinfo{year}{2014}).
\newblock


\bibitem[\protect\citeauthoryear{Sharma, Kiros, and Salakhutdinov}{Sharma
  et~al\mbox{.}}{2015}]%
        {sharma2015action}
\bibfield{author}{\bibinfo{person}{Shikhar Sharma}, \bibinfo{person}{Ryan
  Kiros}, {and} \bibinfo{person}{Ruslan Salakhutdinov}.}
  \bibinfo{year}{2015}\natexlab{}.
\newblock \showarticletitle{Action recognition using visual attention}.
\newblock \bibinfo{journal}{\emph{arXiv:1511.04119}} (\bibinfo{year}{2015}).
\newblock


\bibitem[\protect\citeauthoryear{Sutskever, Vinyals, and Le}{Sutskever
  et~al\mbox{.}}{2014}]%
        {sutskever2014sequence}
\bibfield{author}{\bibinfo{person}{Ilya Sutskever}, \bibinfo{person}{Oriol
  Vinyals}, {and} \bibinfo{person}{Quoc~V Le}.}
  \bibinfo{year}{2014}\natexlab{}.
\newblock \showarticletitle{Sequence to sequence learning with neural
  networks}. In \bibinfo{booktitle}{\emph{NIPS}}.
\newblock


\bibitem[\protect\citeauthoryear{Veeriah, Zhuang, and Qi}{Veeriah
  et~al\mbox{.}}{2015}]%
        {veeriah2015differential}
\bibfield{author}{\bibinfo{person}{Vivek Veeriah}, \bibinfo{person}{Naifan
  Zhuang}, {and} \bibinfo{person}{Guo-Jun Qi}.}
  \bibinfo{year}{2015}\natexlab{}.
\newblock \showarticletitle{Differential recurrent neural networks for action
  recognition}. In \bibinfo{booktitle}{\emph{ICCV}}.
\newblock


\bibitem[\protect\citeauthoryear{Wang, Geng, Ma, Liu, and Yang}{Wang
  et~al\mbox{.}}{2018}]%
        {wang2018crowd}
\bibfield{author}{\bibinfo{person}{Leye Wang}, \bibinfo{person}{Xu Geng},
  \bibinfo{person}{Xiaojuan Ma}, \bibinfo{person}{Feng Liu}, {and}
  \bibinfo{person}{Qiang Yang}.} \bibinfo{year}{2018}\natexlab{}.
\newblock \showarticletitle{Crowd Flow Prediction by Deep Spatio-Temporal
  Transfer Learning}.
\newblock \bibinfo{journal}{\emph{arXiv preprint arXiv:1802.00386}}
  (\bibinfo{year}{2018}).
\newblock


\bibitem[\protect\citeauthoryear{Wang, Chen, Li, Xu, and Lin}{Wang
  et~al\mbox{.}}{2017}]%
        {wang2017multi}
\bibfield{author}{\bibinfo{person}{Zhouxia Wang}, \bibinfo{person}{Tianshui
  Chen}, \bibinfo{person}{Guanbin Li}, \bibinfo{person}{Ruijia Xu}, {and}
  \bibinfo{person}{Liang Lin}.} \bibinfo{year}{2017}\natexlab{}.
\newblock \showarticletitle{Multi-Label Image Recognition by Recurrently
  Discovering Attentional Regions}. In \bibinfo{booktitle}{\emph{CVPR}}.
\newblock


\bibitem[\protect\citeauthoryear{Williams, Durvasula, and Brown}{Williams
  et~al\mbox{.}}{1998}]%
        {williams1998urban}
\bibfield{author}{\bibinfo{person}{Billy Williams}, \bibinfo{person}{Priya
  Durvasula}, {and} \bibinfo{person}{Donald Brown}.}
  \bibinfo{year}{1998}\natexlab{}.
\newblock \showarticletitle{Urban freeway traffic flow prediction: application
  of seasonal autoregressive integrated moving average and exponential
  smoothing models}.
\newblock \bibinfo{journal}{\emph{Transportation Research Record: Journal of
  the Transportation Research Board}} \bibinfo{number}{1644}
  (\bibinfo{year}{1998}), \bibinfo{pages}{132--141}.
\newblock


\bibitem[\protect\citeauthoryear{Wu, Li, Cao, Ji, and Lin}{Wu
  et~al\mbox{.}}{2018}]%
        {wu2018interpretable}
\bibfield{author}{\bibinfo{person}{Xian Wu}, \bibinfo{person}{Guanbin Li},
  \bibinfo{person}{Qingxing Cao}, \bibinfo{person}{Qingge Ji}, {and}
  \bibinfo{person}{Liang Lin}.} \bibinfo{year}{2018}\natexlab{}.
\newblock \showarticletitle{Interpretable Video Captioning via Trajectory
  Structured Localization}. In \bibinfo{booktitle}{\emph{CVPR}}.
  \bibinfo{pages}{6829--6837}.
\newblock


\bibitem[\protect\citeauthoryear{Xingjian, Chen, Wang, Yeung, Wong, and
  Woo}{Xingjian et~al\mbox{.}}{2015}]%
        {xingjian2015convolutional}
\bibfield{author}{\bibinfo{person}{SHI Xingjian}, \bibinfo{person}{Zhourong
  Chen}, \bibinfo{person}{Hao Wang}, \bibinfo{person}{Dit-Yan Yeung},
  \bibinfo{person}{Wai-Kin Wong}, {and} \bibinfo{person}{Wang-chun Woo}.}
  \bibinfo{year}{2015}\natexlab{}.
\newblock \showarticletitle{Convolutional LSTM network: A machine learning
  approach for precipitation nowcasting}. In \bibinfo{booktitle}{\emph{NIPS}}.
\newblock


\bibitem[\protect\citeauthoryear{Xiong, Shi, and Yeung}{Xiong
  et~al\mbox{.}}{2017}]%
        {xiong2017spatiotemporal}
\bibfield{author}{\bibinfo{person}{Feng Xiong}, \bibinfo{person}{Xingjian Shi},
  {and} \bibinfo{person}{Dit-Yan Yeung}.} \bibinfo{year}{2017}\natexlab{}.
\newblock \showarticletitle{Spatiotemporal modeling for crowd counting in
  videos}. In \bibinfo{booktitle}{\emph{ICCV}}.
\newblock


\bibitem[\protect\citeauthoryear{Xu and Saenko}{Xu and Saenko}{2016}]%
        {xu2016ask}
\bibfield{author}{\bibinfo{person}{Huijuan Xu} {and} \bibinfo{person}{Kate
  Saenko}.} \bibinfo{year}{2016}\natexlab{}.
\newblock \showarticletitle{Ask, attend and answer: Exploring question-guided
  spatial attention for visual question answering}. In
  \bibinfo{booktitle}{\emph{ECCV}}.
\newblock


\bibitem[\protect\citeauthoryear{Yao, Wu, Ke, Tang, Jia, Lu, Gong, and Ye}{Yao
  et~al\mbox{.}}{2018}]%
        {yao2018deep}
\bibfield{author}{\bibinfo{person}{Huaxiu Yao}, \bibinfo{person}{Fei Wu},
  \bibinfo{person}{Jintao Ke}, \bibinfo{person}{Xianfeng Tang},
  \bibinfo{person}{Yitian Jia}, \bibinfo{person}{Siyu Lu},
  \bibinfo{person}{Pinghua Gong}, {and} \bibinfo{person}{Jieping Ye}.}
  \bibinfo{year}{2018}\natexlab{}.
\newblock \showarticletitle{Deep multi-view spatial-temporal network for taxi
  demand prediction}.
\newblock \bibinfo{journal}{\emph{arXiv preprint arXiv:1802.08714}}
  (\bibinfo{year}{2018}).
\newblock


\bibitem[\protect\citeauthoryear{Zhang, Zheng, and Qi}{Zhang
  et~al\mbox{.}}{2017b}]%
        {zhang2017deep}
\bibfield{author}{\bibinfo{person}{Junbo Zhang}, \bibinfo{person}{Yu Zheng},
  {and} \bibinfo{person}{Dekang Qi}.} \bibinfo{year}{2017}\natexlab{b}.
\newblock \showarticletitle{Deep Spatio-Temporal Residual Networks for Citywide
  Crowd Flows Prediction.}. In \bibinfo{booktitle}{\emph{AAAI}}.
  \bibinfo{pages}{1655--1661}.
\newblock


\bibitem[\protect\citeauthoryear{Zhang, Zheng, Qi, Li, and Yi}{Zhang
  et~al\mbox{.}}{2016}]%
        {zhang2016dnn}
\bibfield{author}{\bibinfo{person}{Junbo Zhang}, \bibinfo{person}{Yu Zheng},
  \bibinfo{person}{Dekang Qi}, \bibinfo{person}{Ruiyuan Li}, {and}
  \bibinfo{person}{Xiuwen Yi}.} \bibinfo{year}{2016}\natexlab{}.
\newblock \showarticletitle{DNN-based prediction model for spatio-temporal
  data}. In \bibinfo{booktitle}{\emph{ACM SIGSPATIAL}}.
\newblock


\bibitem[\protect\citeauthoryear{Zhang, Lin, Wang, Wang, and Zuo}{Zhang
  et~al\mbox{.}}{2018}]%
        {zhang2018hierarchical}
\bibfield{author}{\bibinfo{person}{Ruimao Zhang}, \bibinfo{person}{Liang Lin},
  \bibinfo{person}{Guangrun Wang}, \bibinfo{person}{Meng Wang}, {and}
  \bibinfo{person}{Wangmeng Zuo}.} \bibinfo{year}{2018}\natexlab{}.
\newblock \showarticletitle{Hierarchical Scene Parsing by Weakly Supervised
  Learning with Image Descriptions}.
\newblock \bibinfo{journal}{\emph{PAMI}} \bibinfo{number}{1}
  (\bibinfo{year}{2018}), \bibinfo{pages}{1--1}.
\newblock


\bibitem[\protect\citeauthoryear{Zhang, Lin, Zhang, Zuo, and Zhang}{Zhang
  et~al\mbox{.}}{2015}]%
        {zhang2015bit}
\bibfield{author}{\bibinfo{person}{Ruimao Zhang}, \bibinfo{person}{Liang Lin},
  \bibinfo{person}{Rui Zhang}, \bibinfo{person}{Wangmeng Zuo}, {and}
  \bibinfo{person}{Lei Zhang}.} \bibinfo{year}{2015}\natexlab{}.
\newblock \showarticletitle{Bit-scalable deep hashing with regularized
  similarity learning for image retrieval and person re-identification}.
\newblock \bibinfo{journal}{\emph{TIP}} \bibinfo{volume}{24},
  \bibinfo{number}{12} (\bibinfo{year}{2015}), \bibinfo{pages}{4766--4779}.
\newblock


\bibitem[\protect\citeauthoryear{Zhang, Wu, Costeira, and Moura}{Zhang
  et~al\mbox{.}}{2017a}]%
        {zhang2017fcn}
\bibfield{author}{\bibinfo{person}{Shanghang Zhang}, \bibinfo{person}{Guanhang
  Wu}, \bibinfo{person}{Jo{\~a}o~P Costeira}, {and}
  \bibinfo{person}{Jos{\'e}~MF Moura}.} \bibinfo{year}{2017}\natexlab{a}.
\newblock \showarticletitle{FCN-rLSTM: Deep Spatio-Temporal Neural Networks for
  Vehicle Counting in City Cameras}.
\newblock \bibinfo{journal}{\emph{arXiv:1707.09476}} (\bibinfo{year}{2017}).
\newblock


\bibitem[\protect\citeauthoryear{Zheng}{Zheng}{2015}]%
        {zheng2015trajectory}
\bibfield{author}{\bibinfo{person}{Yu Zheng}.} \bibinfo{year}{2015}\natexlab{}.
\newblock \showarticletitle{Trajectory data mining: an overview}.
\newblock \bibinfo{journal}{\emph{TIST}} \bibinfo{volume}{6},
  \bibinfo{number}{3} (\bibinfo{year}{2015}), \bibinfo{pages}{29}.
\newblock


\bibitem[\protect\citeauthoryear{Zheng, Capra, Wolfson, and Yang}{Zheng
  et~al\mbox{.}}{2014}]%
        {zheng2014urban}
\bibfield{author}{\bibinfo{person}{Yu Zheng}, \bibinfo{person}{Licia Capra},
  \bibinfo{person}{Ouri Wolfson}, {and} \bibinfo{person}{Hai Yang}.}
  \bibinfo{year}{2014}\natexlab{}.
\newblock \showarticletitle{Urban computing: concepts, methodologies, and
  applications}.
\newblock \bibinfo{journal}{\emph{TIST}} \bibinfo{volume}{5},
  \bibinfo{number}{3} (\bibinfo{year}{2014}), \bibinfo{pages}{38}.
\newblock


\end{thebibliography}

\end{document}